\documentclass[a4paper,twoside]{article}

\usepackage{amsmath}
\usepackage{amssymb}
\usepackage{siunitx}

\usepackage{subcaption}

\usepackage{ifthen}
\usepackage{algorithm}
\usepackage{algpseudocode}

\usepackage{url}

\usepackage{flushend}

\usepackage{tikz}
\usepackage{pgfplots}
\pgfplotsset{compat=1.12}

\usetikzlibrary{3d}
\usetikzlibrary{arrows}

\usepackage{SCITEPRESS}

\usepackage{hyperref}

\begin{document}
\title{Simultaneous Object Detection and Semantic Segmentation}

\author{\authorname{Niels Ole Salscheider\sup{1}}
\affiliation{\sup{1}FZI Research Center for Information Technology, Karlsruhe, Germany}
\email{salscheider@fzi.de}}

\keywords{Autonomous Driving, Computer Vision, Deep Learning, Object Detection, Semantic Segmentation.}

\abstract{
Both object detection in and semantic segmentation of camera images are important tasks for automated vehicles.
Object detection is necessary so that the planning and behavior modules can reason about other road users.
Semantic segmentation provides for example free space information and information about static and dynamic parts of the environment.
There has been a lot of research to solve both tasks using Convolutional Neural Networks.
These approaches give good results but are computationally demanding.
In practice, a compromise has to be found between detection performance, detection quality and the number of tasks.
Otherwise it is not possible to meet the real-time requirements of automated vehicles.
In this work, we propose a neural network architecture to solve both tasks simultaneously.
This architecture was designed to run with around 10\,Hz on 1\,MP images on current hardware.
Our approach achieves a mean IoU of 61.2\% for the semantic segmentation task on the challenging Cityscapes benchmark.
It also achieves an average precision of 69.3\% for cars and 67.7\% for pedestrians on the moderate difficulty level of the KITTI benchmark.
}

\onecolumn \maketitle \normalsize \setcounter{footnote}{0} \vfill

\section{Introduction}
Automated vehicles need a detailed perception of their environment in order to drive safely.
Camera images contain the most information compared to data from other sensors like lidar or radar.
Automated vehicles are therefore usually equipped with cameras and try to make use of this information as much as possible.
However, image processing with neural networks requires a lot of computing power.
In practice this means that compromises are necessary during the design of an automated vehicle:
It is not possible to use huge neural networks due to real-time constraints, even if they are the best-performing ones.
It also is not possible to execute different neural networks for every imaginable task at the same time.

Two common tasks in environment perception from camera images are object detection and semantic segmentation.
Object detection is a corner stone of an automated vehicle.
The behavior generation and planning modules need to reason about objects and their future behavior.
Especially other road users and infrastructure elements like traffic signs and traffic lights are of interest here.
This task can be solved using Convolutional Neural Networks like SSD~\cite{Liu2016}, YOLO~\cite{Redmon2016,Redmon2017,Redmon2018} or Faster R-CNN~\cite{Ren2015}.

But also semantic segmentation plays an important role in an automated vehicle.
It can for example be used to validate that the planned trajectory lies within the drivable area (i.\,e. on the road surface).
If the information about the road geometry is not stored in a map, lanes have to be extracted online from the camera image.
Also this task can be solved using semantic segmentation \cite{Meyer2018}.
Another application for semantic segmentation is mapping and localization:
Only static parts of the environment should be mapped or compared to an existing map.
The segmentation map can be used to mask all dynamic parts.
Popular examples of neural networks for semantic segmentation include DeepLab v3 \cite{Chen2018} and PSPNet \cite{Zhao2017}.

Both object detection and semantic segmentation have been extensively researched.
While current approaches do not yet reach human-level performance they are getting close and their accuracy continues to increase.
They also provide valuable information for automated vehicles.
It is therefore important to run these two tasks in parallel while satisfying all real-time constraints.

In this work we present a neural network architecture that solves these tasks jointly.
It was designed to achieve a framerate of around 10\,Hz on 1\,MP images on current hardware.

\section{Related Work}

There is a lot of research on different approaches to object detection and semantic segmentation.
The following section can only give an overview over the most important and recent approaches.

Object detection approaches can be separated into proposal-based ones and proposal-free ones.
A well-known proposal-based approach is Faster R-CNN \cite{Ren2015} and its predecessors R-CNN~\cite{Girshick2014} and Fast R-CNN~\cite{Girshick2015}.
These approaches first generate object proposals and then predict for each proposal if it is an object or not.
Faster R-CNN generates these proposals using a Region Proposal Network while its predecessors use Selective Search~\cite{Sande2011} to do so.
In the case of Faster R-CNN, this Region Proposal Network is a Convolutional Neural Network that takes the whole image as input.
For each proposal, the CNN features of the proposed Region of Interest are reshaped using a pooling layer and then fed into two heads.
One classifies the proposal and decides if it is an object or not.
The other regresses the bounding box.

Proposal-based approaches give good results but they are usually slower than proposal-free approaches.
One notable example of the latter category is SSD~\cite{Liu2016}.
It's design is based on the idea of anchor boxes.
The output space is discretized into a fixed set of anchor boxes with different scales and aspect ratios for each feature map location.
The authors use feature maps with different resolutions to capture objects of different sizes.
During inference, the network predicts scores for each anchor box that indicate if the anchor box contains an object of a specific class.
It also gives a regression of the bounding box offset relative to the anchor box.
Finally, non-maxima suppression is applied to all predicted bounding boxes.

YOLO~\cite{Redmon2016} splits the image into a grid and performs object detection in each cell.
For each grid cell the network outputs a fixed number of bounding boxes with class probabilities and bounding box regression.
For the successors \cite{Redmon2017,Redmon2018}, the authors remove the fully connected layers for direct box regression and replace them by anchor boxes.

Another proposal-free approach is RetinaNet~\cite{Lin2017}.
It draws from the ideas of other detection approaches to build a simple model.
The authors propose a new loss function called  Focal Loss that can deal with the high foreground-background imbalance without sampling.
With this, the comparatively simple model can achieve state-of-the-art performance.

Pixel-wise semantic segmentation with CNNs became popular when FCN \cite{Long2015} started to use fully convolutional networks.
SegNet \cite{Badrinarayanan2017} then introduced an encoder-decoder structure to produce high-resolution segmentation maps.
Popular examples that achieve state-of-the-art results include PSPNet \cite{Zhao2017} and DeepLab v3 \cite{Chen2018}.
Both employ a form of spatial pyramid pooling to capture context at different scales.

In recent years, multi-task learning has gained more popularity.
Solving multiple tasks at once does not only reduce the computational demand compared to solving them sequentially.
The different training objectives can also act as regularizers that make the model generalize better.
The model is encouraged to learn more generic features that help to solve all tasks \cite{Baxter2000}.

There has also been work on joint learning of semantic segmentation and object detection.
In \cite{Uhrig2016}, the authors describe an approach to instance segmentation using multi-task learning.
For each pixel they predict the class label, depth and the direction to the corresponding instance center using a single neural network.
They then decode the instance masks from this representation.

Recently, two approaches that are similar to ours have been published \cite{Teichmann2018,Sistu2019}.
Both learn segmentation and object detection in a multi-task setting.
However, \cite{Teichmann2018} only predicts a road segmentation.
The network structure proposed in \cite{Sistu2019} is considerably simpler and smaller.
As a consequence, the inference time of their neural network is much lower but the accuracy is also notably worse.

\section{Approach}

In the following we will first present the design of our proposed neural network.
In the next sub-section we will give the training details.
The code that was used to perform the experiments in this work is available as open source software\footnote{\url{https://github.com/fzi-forschungszentrum-informatik/NNAD/tree/icpram2020}}.

\subsection{Network Design}

\begin{figure}
 \begin{center}
  \includegraphics[scale=0.8]{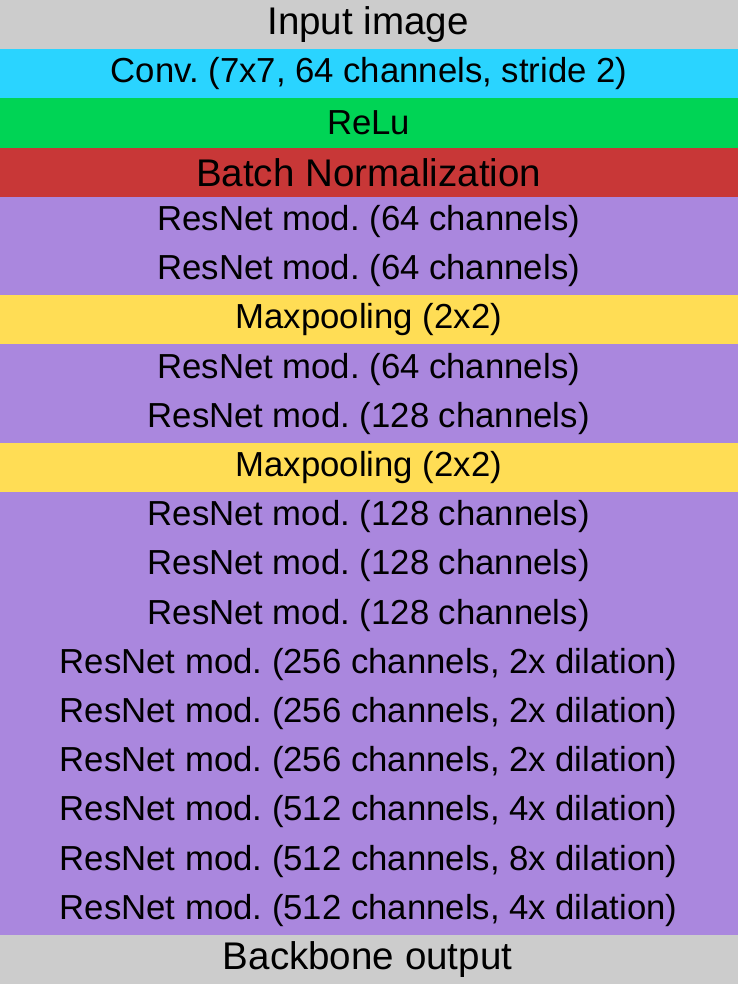}
 \end{center}
 \caption{Structure of the proposed backbone for simultaneous semantic segmentation and object detection.
  The ResNet modules have the same structure that is proposed in \cite{He2016} but use depthwise separable convolutions to reduce the computational demand.}
 \label{fig:backbone}
\end{figure}

In this work we propose a network structure for simultaneous semantic segmentation and object detection.
The backbone of our model is based on ResNet-38 \cite{Wu2016}.
The structure of the backbone is visualized in Figure~\ref{fig:backbone}.
The ResNet modules have the same structure that is proposed in \cite{He2016}.
But like Xception \cite{Chollet2017}, we use depthwise separable convolutions to reduce the computational demand.

The path for semantic segmentation has a convolutional encoder-decoder structure.
In the first layers of the encoder, the data tensor is sampled down by a factor of 8:
The first convolution has a stride of two and there are two maxpooling layers that both downsample by a factor of two.
This reduction of resolution is necessary to decrease the computational demand of the network.
But after this reduction we use dilated convolutions as proposed in \cite{Chen2017}.
This increases the receptive field of the convolutions while preserving spatial details and while keeping the number of learnable parameters constant.

\begin{figure}
 \begin{center}
  \includegraphics[scale=0.8]{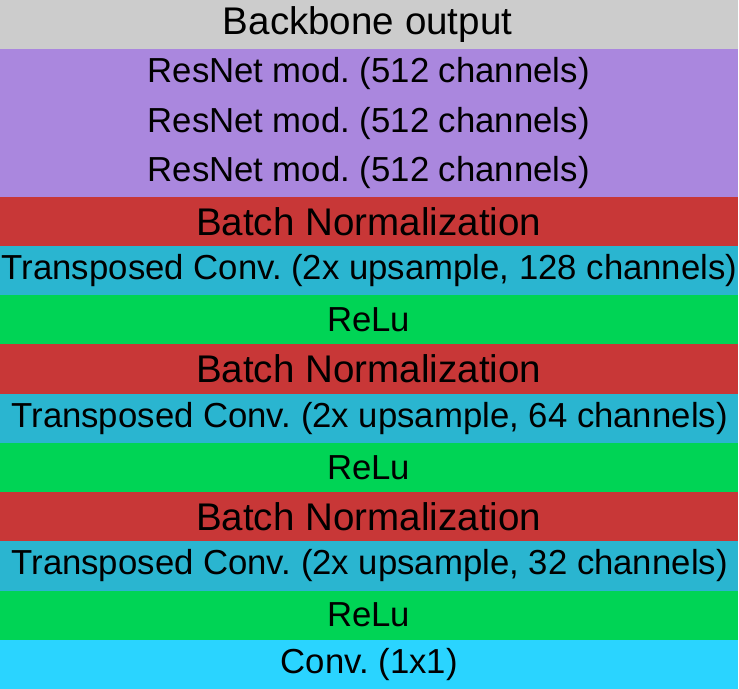}
 \end{center}
 \caption{Structure of the network head for semantic segmentation.
 The data tensor is upsampled by a factor of 8 to compensate the downsampling in the backbone.}
 \label{fig:segmentation_head}
\end{figure}

The output of the backbone is then fed into multiple network heads.
One head is the semantic segmentation head.
It is visualized in Figure~\ref{fig:segmentation_head}.
After three more ResNet modules the data tensor is upsampled again so that the final segmentation map has the same resolution as the input image.
This is done using three transposed convolutions that each learn to upsample by a factor of 2.
An alternative would be to upsample by a factor of 8 with just one transposed convolution.
But this way we can gradually reduce the number of channels while increasing the spatial resolution.
The final convolution layer then reduces the number of channels to the number of classes.
During training, a softmax function is applied to its output and it is trained using cross-entropy loss.

\begin{figure}
 \begin{center}
  \includegraphics[scale=0.8]{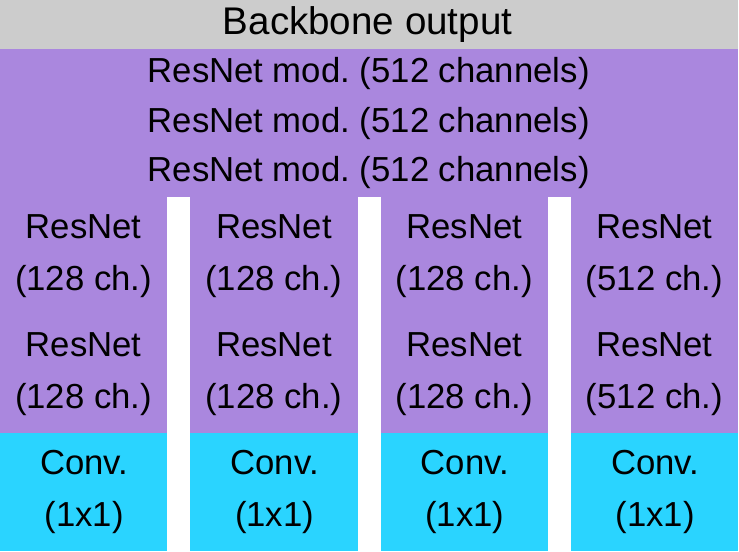}
 \end{center}
 \caption{Structure of the network head for object detection.
 The different outputs are an objectness score, a class score, bounding box parameters regression and a feature embedding per anchor box.}
 \label{fig:box_head}
\end{figure}

The second head of our proposed model is the object detection head.
It also takes the output of the backbone as input and applies three more shared ResNet modules.
The output of the last shared ResNet module is fed into four sub-networks of identical structure but with a different number of channels.
Each consists of two more ResNet blocks and a final convolutional layer to adjust the number of channels for the final task.

The first sub-network solves a binary classification problem.
It predicts whether the corresponding anchor box contains a relevant object or not.
During training, a softmax function is applied to its output.
Like RetinaNet \cite{Lin2017} we use Focal Loss to train this output.
We chose $\alpha = 1.0$ and $\gamma = 2.0$ as parameters.

The second sub-network also solves a classification problem.
For all anchor boxes that contain a relevant object it predicts its class.
Like for the semantic segmentation head, a softmax function is applied to its output and it is trained using cross-entropy loss for all active anchor boxes.

\begin{table}
 \begin{tabular}{| l | l |}
  \hline
  Anchor box       & 0.25, 0.5, 1.0, 2.0, 4.0 \\
  ratios (w/h)     & \\
  \hline
  Anchor box       & 32, 48, 64, 96, 128, 192, \\
  areas (in pixel) & 256, 384, 512, 768, 1024, \\
                   & 1536, 2048, 3072, 4096, \\
                   & 6144, 8192, 12288, 16384, \\
                   & 24576, 32768, 49152, 65536, \\
                   & 98304, 131072, 196608, \\
                   & 262144, 393216, 524288 \\
  \hline
 \end{tabular}

 \caption{The parameters of our anchor boxes.
 At each location of the feature map we generate anchor boxes for all possible combinations of box ratio and box area.}
 \label{tab:anchor_parameters}
\end{table}

The third sub-network gives the regression output of the bounding box parameters.
These parameters are the same as in R-CNN \cite{Girshick2014}.
The parameters for the used anchor boxes can be found in Table~\ref{tab:anchor_parameters}.
We generate anchor boxes at each location of the feature map for all possible combinations of box ratio and box area.
Since we only downsample by a factor of 8 in the encoder to preserve spatial details for the semantic segmentation task we do not have low-resolution feature maps.
In contrast to RetinaNet \cite{Lin2017} we therefore only predict objects on one feature map.
In order to still be able to detect objects of different sizes, we generate more anchor boxes with different scales.
We train the output with smooth L1 loss for all active anchor boxes.

The fourth sub-network is optional.
If desired it can be used to learn a feature embedding for each detected object.
We include it here because it is useful for some applications and we will use it in future work.
The feature embedding is trained using contrastive loss \cite{Hadsell2006} for all active anchor boxes.
Since we train the network on single images and not sequences all training examples come from one image.
We use all anchor boxes that correspond to the same object as positive examples and all that correspond to other objects as negative examples.

\subsection{Training Procedure}

We train our model on the Cityscapes dataset \cite{Cordts2016}.
It contains 5\,000 finely annotated and 20\,000 coarsely annotated training images.
In order to train the object detection head we extract bounding boxes from the available instance labels.
We do this by taking the minimum and maximum of the x- and y-coordinates of the instance polygons.

Our model is trained with the Adam optimizer \cite{Kingma2014}.
Like \cite{Zhao2017}, we use a polynomial decay learning rate schedule of the following form:
\begin{equation*} lr(iter) = base\_lr \cdot \left(1.0 - \frac{iter}{max\_iter}\right)^{0.9} \end{equation*}
We use a batch size of 8 and a base learning rate of 0.001 and run the training for 300\,000 iterations.

We use the approach proposed in \cite{Kendall2017} to weight the different losses of all tasks.

\subsection{Bounding Box Target Generation}
During training the target output of the neural network for the bounding boxes is generated from the ground truth bounding box list.
Our procedure for this is as follows:
We initialize all anchor boxes as being ``inactive'' (i.\,e. not corresponding to an object).
Then for each ground truth bounding box we calculate the intersection over union with all anchor boxes.
If the IoU value is higher than 0.5 we set the anchor box to being ``active'' and assign the class and regression parameters.
If the IoU value is between 0.4 and 0.5 we set the anchor box to ``don't care''.
This ensures that we do not get high losses and oscillating behavior for anchor boxes right at the threshold.

\begin{figure}
 \begin{center}
  \includegraphics[width=0.8\linewidth]{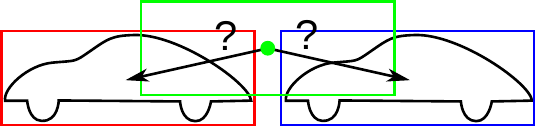}
 \end{center}
 \caption{The green anchor box has high overlap with two objects and would be ``active'' for both.
  Since both would compete in the training objective for the box regression target the learned displacement would average out.
  Therefore the decoded box would not align with any of the true bounding boxes (red and blue) but would be in the middle.
  We avoid this problem by setting these problematic anchor boxes to ``inactive'' during training.}
 \label{fig:assignment_possibilities}
\end{figure}

There are a few corner cases that we also take into account:
If parts of the anchor box are outside of the image but it contains an object we set it to ``don't care''.
We do this to avoid conflicting objectives for the box regression task.
If a ground truth bounding box was not assigned to any anchor box (because there is none with an IoU higher than 0.5) but there is an anchor box with IoU higher than 0.4 we assign it to this.
This helps to also detect small objects that fall between the grid.
In case an anchor box can be assigned to multiple ground truth bounding boxes we choose the one with the highest IoU.
But if the absolute difference between the highest and the second highest IoU value is less than 0.2 and both are higher than 0.4 we set the anchor box to ``inactive''.
This helps to ensure that objects are clearly separated.
We found that otherwise the regression output is just the average of the displacements for all overlapping objects.
This means that the decoded bounding box ends up being between the adjacent objects.
It then is too far away from all of them to be suppressed by the non-maxima suppression.
The problem is illustrated in Figure~\ref{fig:assignment_possibilities}.

\section{Evaluation}

\begin{table}
 \begin{center}
  \begin{subtable}{0.7\linewidth}
   \begin{tabular}{l | c | c}
Class          & IoU  & iIoU    \\
\hline
\hline
road          & 0.963 &     -   \\
sidewalk      & 0.736 &     -   \\
building      & 0.868 &     -   \\
wall          & 0.347 &     -   \\
fence         & 0.385 &     -   \\
pole          & 0.459 &     -   \\
traffic light & 0.498 &     -   \\
traffic sign  & 0.625 &     -   \\
vegetation    & 0.886 &     -   \\
terrain       & 0.517 &     -   \\
sky           & 0.905 &     -   \\
person        & 0.711 &   0.518 \\
rider         & 0.456 &   0.310 \\
car           & 0.903 &   0.809 \\
truck         & 0.381 &   0.212 \\
bus           & 0.614 &   0.381 \\
train         & 0.347 &   0.176 \\
motorcycle    & 0.350 &   0.261 \\
bicycle       & 0.653 &   0.462 \\
\hline
Average       & 0.612 &   0.395 \\
\hline
\hline
   \end{tabular}
  \caption{Per-class results.}
  \end{subtable}

\vspace{1em}
  
  \begin{subtable}{0.7\linewidth}
   \begin{tabular}{l | c | c}
Category      & IoU  & iIoU     \\
\hline
\hline
flat          & 0.977 &     -   \\
construction  & 0.872 &     -   \\
object        & 0.530 &     -   \\
nature        & 0.887 &     -   \\
sky           & 0.905 &     -   \\
human         & 0.721 &   0.543 \\
vehicle       & 0.881 &   0.774 \\
\hline
Average       & 0.825 &   0.658 \\
   \end{tabular}
  \caption{Per-category results.}
  \end{subtable}
 \end{center}

 \caption{Results of the semantic segmentation on the Cityscapes validation dataset.
  The results have been computed with the official evaluation script.
  The IoU metric is the intersection-over-union metric used by PASCAL VOC \cite{Everingham2015}.
  The iIoU metric is computed by weighting each pixel with the ratio of the average instance size and the size of the ground truth instance size.
  }
 \label{tab:segmentation_results}
\end{table}

We train and evaluate our proposed neural network architecture on the Cityscapes benchmark \cite{Cordts2016}.
The results for the semantic segmentation task can be found in Table~\ref{tab:segmentation_results}.

\begin{figure}
 \begin{subfigure}{0.49\linewidth}
  \begin{center}
   \includegraphics[width=\linewidth]{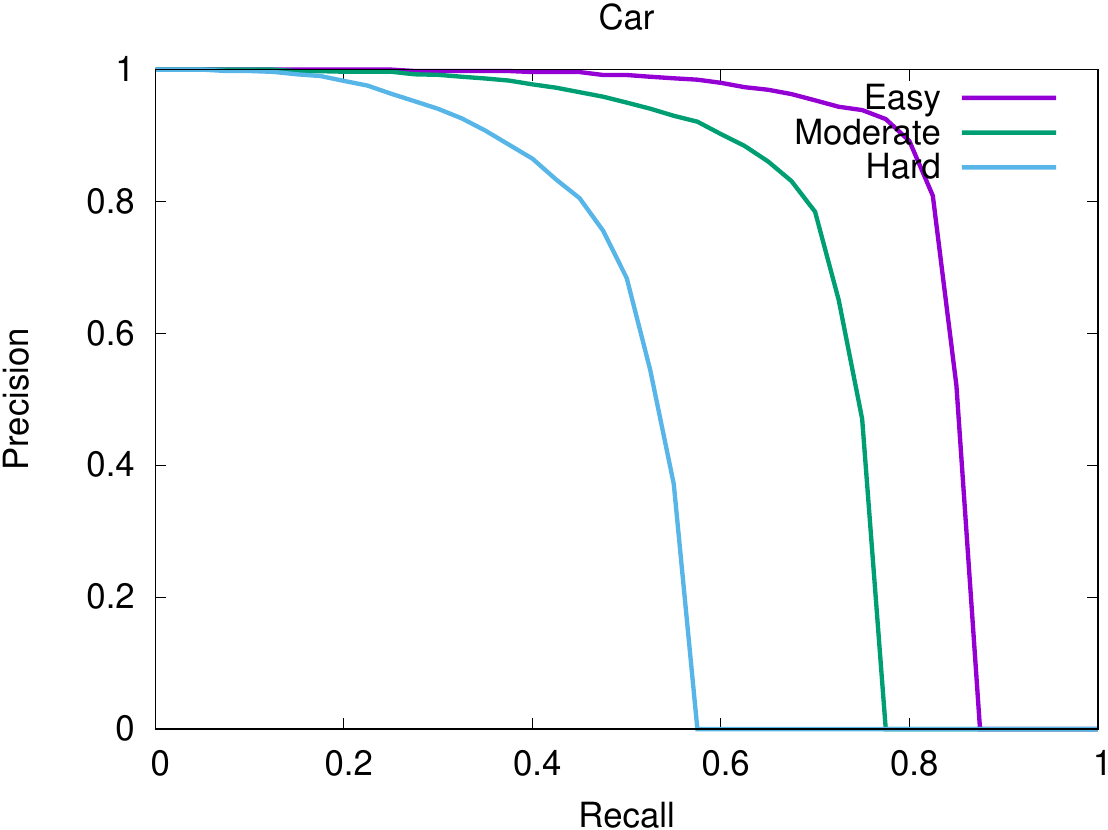}
  \end{center}
  \caption{Car class.}
 \end{subfigure}
 \begin{subfigure}{0.49\linewidth}
  \begin{center}
   \includegraphics[width=\linewidth]{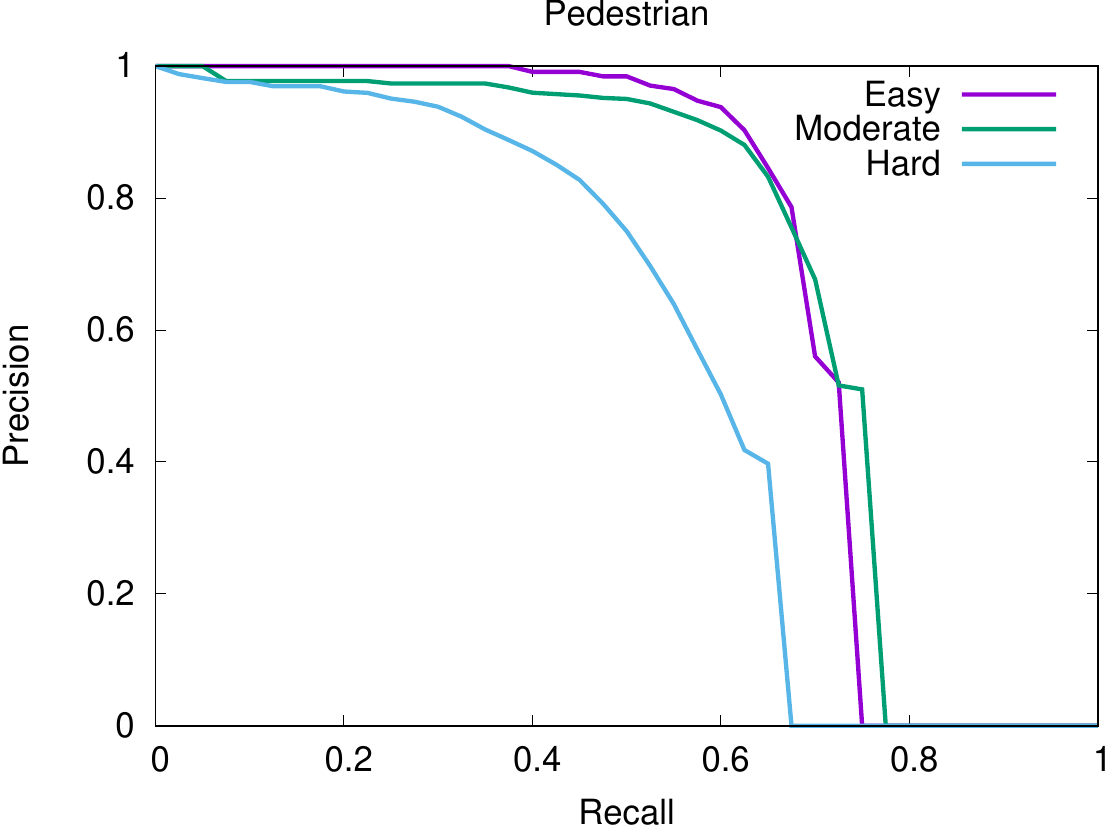}
  \end{center}
  \caption{Pedestrian class.}
 \end{subfigure}
 \\[2em]
 \begin{subtable}{\linewidth}
  \begin{center}
   \begin{tabular}{ l || l | l | l }
    Class & Easy & Moderate & Hard \\
    \hline
    \hline
    Car        & 84.5\% & 72.3\% & 50.9\% \\
    Pedestrian & 70.9\% & 70.8\% & 56.5\% \\
   \end{tabular}
   \caption{Average Precision values.}
  \end{center}
 \end{subtable}

 \caption{Results of the object detection on the Cityscapes validation dataset.
 It uses the KITTI evaluation tool but with an adjusted definition of the difficulty levels that is better suited for Cityscapes.}
 \label{fig:results_object_cs}
\end{figure}

We evaluate the performance of our bounding box detector for the ``car'' and ``pedestrian'' classes with the official evaluation tool of the KITTI benchmark \cite{Geiger2012}.
The results can be found in Figure~\ref{fig:results_object_cs}.
Since we train the model with the Cityscapes dataset we also want to evaluate on this dataset.
It however lacks annotations for the level of truncation and occlusion.
We therefore ignore these values during the evaluation, making the task harder.
But since the images from the Cityscapes dataset have a higher resolution than the ones from the KITTI dataset we adjusted the size limits for the evaluation difficulty levels.
Here, we require a minimum width and height of 10 for the ``hard'' difficulty level, 50 for the ``moderate'' difficulty level and 100 for the ``easy'' difficulty level.
Because of these differences the results are not comparable with the ones achieved on the KITTI dataset.
But we hope that these are useful values that can be used in future work for comparisons on the Cityscapes dataset.

Especially the ``car'' class has many examples with high occlusion levels in the Cityscapes dataset.
This explains the low recall in the ``hard'' difficulty level for this class.

\begin{figure}
 \begin{subfigure}{0.49\linewidth}
  \begin{center}
   \includegraphics[width=\linewidth]{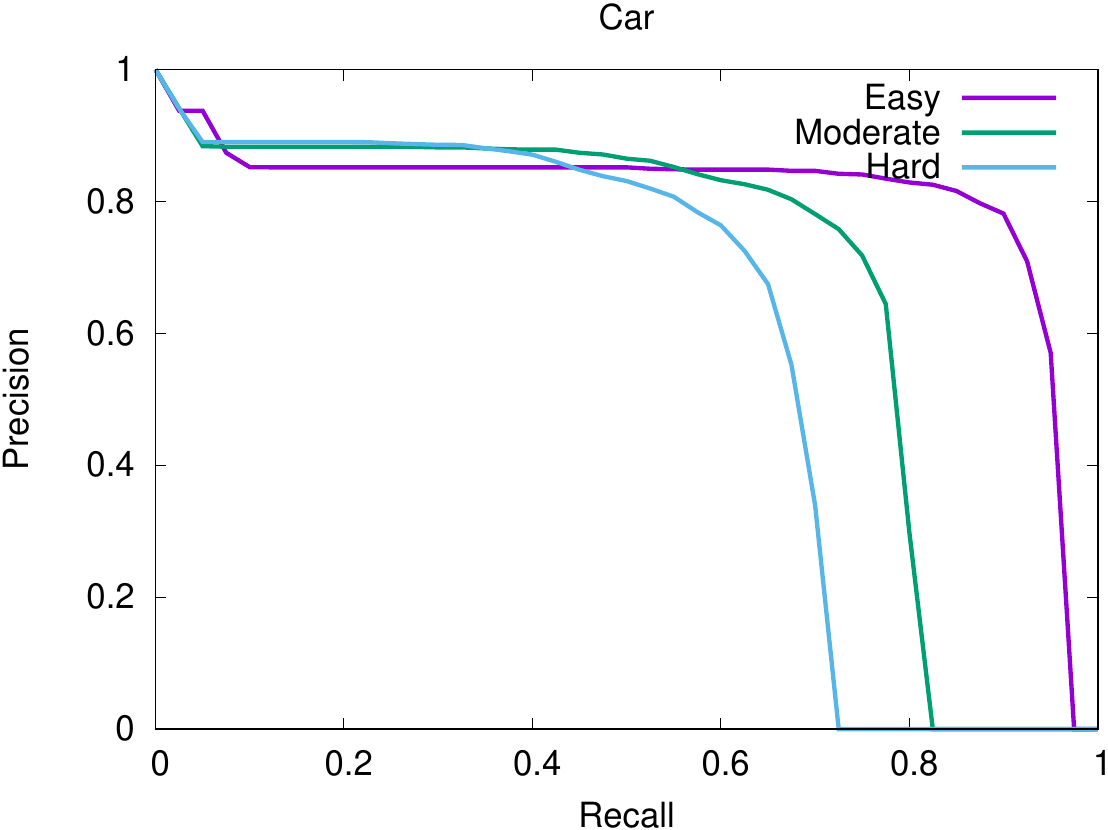}
  \end{center}
  \caption{Car class.}
 \end{subfigure}
 \begin{subfigure}{0.49\linewidth}
  \begin{center}
   \includegraphics[width=\linewidth]{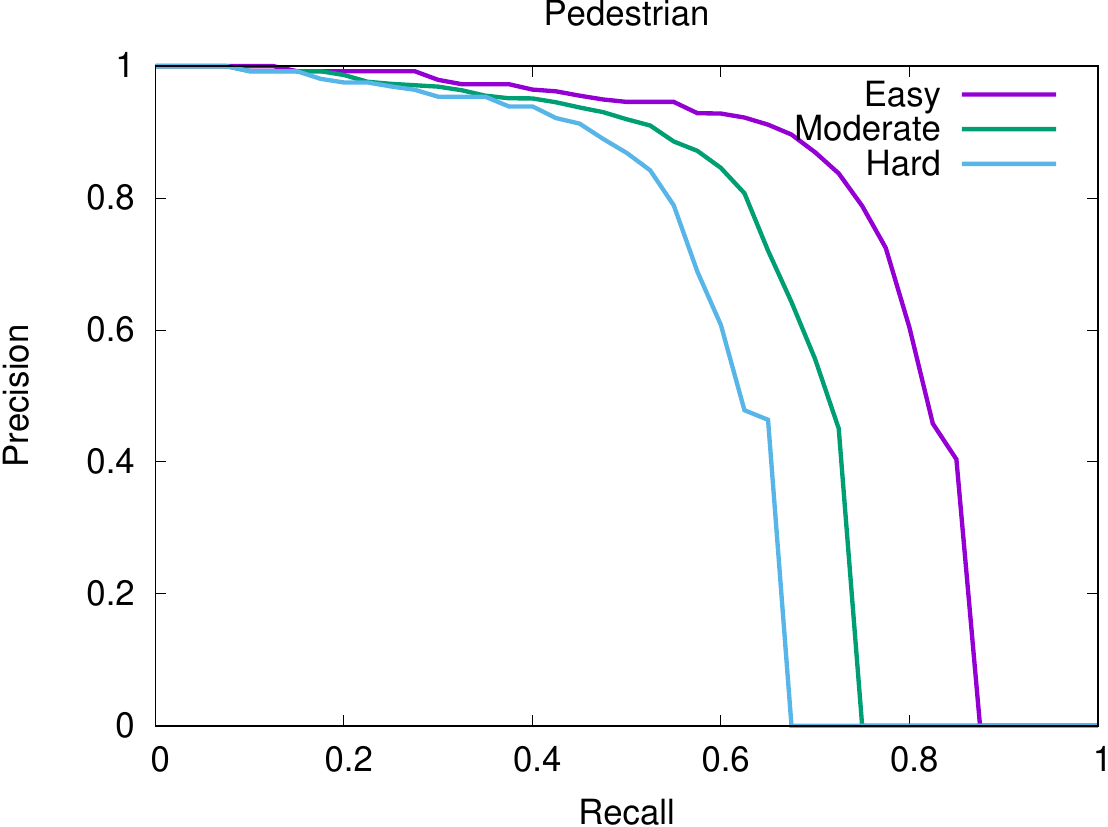}
  \end{center}
  \caption{Pedestrian class.}
 \end{subfigure}
 \\[2em]
 \begin{subtable}{\linewidth}
  \begin{center}
   \begin{tabular}{ l || l | l | l }
    Class & Easy & Moderate & Hard \\
    \hline
    \hline
    Car        & 82.1\% & 69.3\% & 60.2\% \\
    Pedestrian & 79.5\% & 67.7\% & 60.1\% \\
   \end{tabular}
   \caption{Average Precision values.}
  \end{center}
 \end{subtable}

 \caption{Results of the object detection on the KITTI validation dataset.
 Here we follow the official definition of the difficulty levels.}
 \label{fig:results_object_kitti}
\end{figure}

We also evaluate the object detection performance on the KITTI dataset.
Here we follow the official definitions of the difficulty levels from KITTI.
The results can be found in Figure~\ref{fig:results_object_kitti}.
We randomly select 5\,930 images from the KITTI training dataset to form a validation dataset.
Then we mix the remaining images with the Cityscapes training images and fine-tun our model with that.
One problem is that the generated bounding box labels from the Cityscapes dataset and the labels from the KITTI dataset are not consistent:
While the generated bounding boxes cover only the visible parts of each object the labels from the KITTI dataset cover the projection of the whole object.
This explains the drop in precision for the ``car'' class even for detection thresholds with low recall.
It also means that we observe lower recall at detection thresholds with low precision.
Another issue is that the images from the KITTI dataset have a lower resolution while we tuned our model for the notably higher resolution of the Cityscapes dataset.
These results are therefore not directly comparable with a detector that is only trained on the KITTI dataset.
They however give a lower bound for the expected performance.

Our proposed architecture does not reach the level of accuracy that is achieved by the best-performing approaches on the Cityscapes and KITTI leaderboards at the time of writing.
It however gives good accuracy while meeting the desired computation time constraints.
The inference time of the proposed convolutional neural network for images at the desired resolution of 1\,MP is 102\,ms on an Nvidia Titan V GPU.
We measured this time using TensorFlow 1.13 and Nvidia TensorRT 5.1.2.2 RC at a precision of 16\,bit.

\section{Conclusion and Outlook}

We demonstrate that two important vision tasks for automated vehicles (semantic segmentation and object detection) can be learned jointly by a single CNN.
We present a suitable neural network architecture for this which takes the needs of both tasks into account.
It does not achieve the level of accuracy that the the best-performing models offer today.
However, it gives good accuracy while meeting the run-time constraints of the application:
Our approach achieves the design goal of a framerate of 10\,Hz on 1\,MP images.

We currently use the presented approach in our research vehicle.
The semantic segmentation gives information about the static and dynamic parts of the world.
This information is useful for mapping where we only want to map the static parts.
It can also be used for mapless driving or freespace validation.
The object detections are fused with detections from other sensors and then used by the behavior and trajectory planning modules.

We will focus on further reducing the run-time of our object detection head without losing accuracy.
In future work we will present a tracking approach for road users that is based on this work.

\bibliographystyle{./apalike}
{\small
\bibliography{./mybibliography}}

\end{document}